\documentclass[11pt,a4paper]{article}
\usepackage[hyperref]{emnlp-ijcnlp-2019}
\usepackage{times}
\usepackage{latexsym}

\usepackage{url}

\usepackage{times}
\usepackage{latexsym}
\usepackage{amsmath,amsfonts,amssymb}
\usepackage{subcaption}

\usepackage{booktabs}
\usepackage{multirow}
\usepackage{graphicx}
\usepackage{subfiles}
\usepackage{todonotes}
\usepackage{xspace}

\aclfinalcopy 
\def\aclpaperid{75} 

\newcommand{\en}{\textsc{en}\xspace}
\newcommand{\es}{\textsc{es}\xspace}
\newcommand{\de}{\textsc{de}\xspace}
\newcommand{\ita}{\textsc{it}\xspace} 
\newcommand{\fr}{\textsc{fr}\xspace}
\newcommand{\ru}{\textsc{ru}\xspace}
\newcommand{\zh}{\textsc{zh}\xspace}
\newcommand{\ja}{\textsc{ja}\xspace}

\newcommand{\B}[1]{%
    \textbf{#1} 
}

\newcommand\tab[1][1cm]{\hspace*{#1}}

\title{MultiFiT: Efficient Multi-lingual Language Model Fine-tuning}

\author{Julian Eisenschlos\textsuperscript{1}$^\dagger$\footnotemark $\:$ \tab Sebastian Ruder\textsuperscript{2,3}$^\ddagger$\footnotemark[1]$\:$ \tab Piotr Czapla\textsuperscript{4}\footnotemark[1]$\:$\\\textbf{Marcin Kardas\textsuperscript{4}\footnotemark[1]$\:$ \tab Sylvain Gugger\textsuperscript{5} \tab Jeremy Howard\textsuperscript{5,6}} \\
  \textsuperscript{1}ASAPP, Inc. \tab  \textsuperscript{2}National University of Ireland \tab
  \textsuperscript{3}Aylien Ltd., Dublin \\
  \textsuperscript{4}n-waves, Wroc{\l}aw \tab 
  \textsuperscript{5}fast.ai \tab
  \textsuperscript{6}University of San Francisco \\
}

\date{}

\begin{document}
\maketitle
\begin{abstract}
Pretrained language models are promising particularly for low-resource languages as they only require unlabelled data. However, training existing models requires huge amounts of compute,
while pretrained cross-lingual models often underperform on low-resource languages.
We propose Multi-lingual language model Fine-Tuning (MultiFiT) to enable practitioners to train and fine-tune language models efficiently in their own language. In addition, we propose a zero-shot method using an existing pretrained cross-lingual model. We evaluate our methods on two widely used cross-lingual classification datasets where they outperform models pretrained on orders of magnitude more data and compute. We release all models and code\footnote{\url{http://nlp.fast.ai}}.
\end{abstract}

\newenvironment{starfootnotes}
  {\par\edef\savedfootnotenumber{\number\value{footnote}}
  \renewcommand{\thefootnote}{$\star$} 
  \setcounter{footnote}{0}}
  {\par\setcounter{footnote}{\savedfootnotenumber}}
  
\begin{starfootnotes}
\footnotetext{The first four authors contributed equally.}
\end{starfootnotes}

\newcommand{\customfootnotetext}[2]{{
  \renewcommand{\thefootnote}{#1}
  \footnotetext[0]{#2}}}

\customfootnotetext{$\dagger$}{Corresponding author: {\tt eisenjulian@gmail.com}}

\customfootnotetext{$\ddagger$}{Sebastian is now affiliated with DeepMind.}

\section{Introduction}

Pretrained language models (LMs) have shown striking improvements on a range of natural language processing (NLP) tasks \cite{Peters2018,Howard2018,Devlin2018}. These models only require unlabelled data for training and are thus particularly useful in scenarios where labelled data is scarce.
As much of NLP research has focused on the English language, the larger promise of these models is to bridge the digital language divide\footnote{\url{http://labs.theguardian.com/digital-language-divide/}} and enable the application of NLP methods to many of the world's other 6,000 languages where labelled data is less plentiful.

Recently, cross-lingual extensions of these LMs have been proposed that train on multiple languages jointly \cite{Artetxe2018e,Lample2019}. These models are able to perform zero-shot learning, only requiring labelled data in the source language. However, source data in another language may often not be available, whereas obtaining a small number of labels is typically straightforward.

Furthermore such models have several downsides: a) some variants rely on large amounts of parallel data, which may not be available for truly low-resource languages; b) they require a huge amount of compute for training\footnote{The training cost is amortized over time as pretraining only needs to be performed once and fine-tuning is much cheaper. However, if a model needs to be applied to a new language or a domain not covered by the model, a new model needs to be trained from scratch.}; and c) cross-lingual models underperform on low-resource languages---precisely the setting where they would be most useful.
We are aware of two possible reasons for this: 1) Languages that are less frequently seen during training are underrepresented in the embedding space.\footnote{This is similar to how word embeddings are known to underperform on low-frequency tokens \cite{Gong2018}.} 2) Infrequent scripts are over-segmented in the shared word piece vocabulary \cite{Wang2019}. 

In this work, we show that small monolingual LMs are able to outperform expensive cross-lingual models both in the zero-shot and the supervised setting. We propose Multi-lingual language model Fine-tuning (MultiFit) to enable practitioners to train and fine-tune language models efficiently.\footnote{We use `multilingual' as referring to training \emph{independent} models in multiple languages. We use `cross-lingual` to refer to training a \emph{joint} model across multiple languages.} Our model combines universal language model fine-tuning \cite[ULMFiT;][]{Howard2018} with the quasi-recurrent neural network \cite[QRNN;][]{Bradbury2017} and subword tokenization \cite{Kudo2018} and can be pretrained on a single Tesla V100 GPU in a few hours. In addition, we propose to use a pretrained cross-lingual model's predictions as pseudo labels to adapt the monolingual language model to the zero-shot setting. We evaluate our models on two widely used cross-lingual classification datasets, MLDoc \cite{Schwenk2018} and CLS \cite{Prettenhofer2010a} where we outperform the state-of-the-art zero-shot model LASER \cite{Artetxe2018e} and multi-lingual BERT \cite{Devlin2018} in the supervised setting---even without any pretraining. In the zero-shot setting, we outperform both models using pseudo labels---and report significantly higher performance with as little as 100 examples. We finally show that information from monolingual and cross-lingual language models is complementary and that pretraining makes models robust to noise.

\section{Related work}

\paragraph{Pretrained language models} Pretrained language models based on an LSTM \cite{Peters2018,Howard2018} and a Transformer \cite{Radford2018,Devlin2018} have been proposed.
Recent work \cite{Peters2018a} suggests that---all else being equal---an LSTM outperforms the Transformer in terms of downstream performance. For this reason, we use a variant of the LSTM as our language model.

\paragraph{Cross-lingual pretrained language models} The multi-lingual BERT model is pretrained on the Wikipedias of 104 languages using a shared word piece vocabulary. LASER \cite{Artetxe2018e} is trained on parallel data of 93 languages with a shared BPE vocabulary. XLM \cite{Lample2019} additionally pretrains BERT with parallel data. These models enable zero-shot transfer, but achieve lower results than monolingual models. In contrast, we focus on making the training of monolingual language models more efficient in a multi-lingual context. 
Concurrent work \cite{Mulcaire2019} pretrains on English and another language, but shows that cross-lingual pretraining only helps sometimes.

\paragraph{Multi-lingual language modeling} Training language models in non-English languages has only recently received some attention. \citet{Kawakami2017} evaluate on seven languages. \citet{Cotterell2018} study 21 languages. \citet{Gerz2018} create datasets for 50 languages. All of these studies, however, only create small datasets, which are inadequate for pretraining language models. In contrast, we are among the first to report the performance of monolingual language models on downstream tasks in multiple languages.

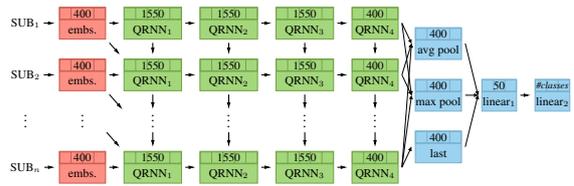
\begin{figure}
\centering
\begin{tikzpicture}[scale=0.5, every node/.style={scale=0.5}]
	\tikzset{>={Latex[width=0.5mm,length=1mm]}}
	  \definecolor{rnn-inner}{HTML}{a4d77c}
  \definecolor{rnn-outer}{HTML}{74a74c}
  \definecolor{emb-inner}{HTML}{fd8f83}
  \definecolor{emb-outer}{HTML}{cd5f53}
  \definecolor{out-inner}{HTML}{9bd2f0}
  \definecolor{out-outer}{HTML}{7bb2d0}

  \tikzstyle{rnn} = [draw=rnn-outer,fill=rnn-inner,inner sep=0.1cm,outer sep=0.1cm]
  \tikzstyle{dots} = [draw=none,fill=none,outer sep=0.1cm]
  \tikzstyle{dense} = [rectangle,inner sep=0,outer sep=0.1cm]
  \tikzstyle{embs} = [fill=emb-inner,draw=emb-outer,outer sep=0.1cm]
  \tikzstyle{rnnout} = [fill=out-inner,draw=out-outer,outer sep=0.1cm]
  \tikzstyle{lbl} = [outer sep=0cm,inner sep=0.1cm]
  
  \def\sblock#1#2#3#4#5{\node[#1,rectangle split,rectangle split parts=2,inner sep=0cm] (#5) {\nodepart{one}\mygrid{#1,inner sep=0}{#3}{#4} \nodepart{two} \mystrut \small #2};}
  
  \def\qrnn#1#2#3#4#5{\node[rnn,rectangle split,rectangle split parts=2,inner sep=0cm,outer xsep=#5] (r#1#2) {\nodepart{one}\mygrid{rnn,inner sep=0}{#3}{#4} \nodepart{two} \mystrut \small QRNN$_#1$};}
  \def\pool#1#2#3#4{\node[rnnout,rectangle split,rectangle split parts=2,inner sep=0cm] (#1) {\nodepart{one}\mygrid{rnnout,inner sep=0}{#3}{#4} \nodepart{two} \mystrut \small #2};}
  \def\mystrut{\vrule height 0.35cm depth 0.15cm width 0pt}
  \def\row#1#2#3#4{

	\qrnn{#1}{0}{#2}{#3}{#4} \&
	\qrnn{#1}{1}{#2}{#3}{#4} \&
	\node (r#12) {$\cdots$}; \&
	\qrnn{#1}{3}{#2}{#3}{#4}
  }

  \def\mygrid#1#2#3{\tikz{\draw[step=0.3,#1] (0,0)  grid (0.3*#2,0.3); \node[rectangle] at (0.15*#2,0.15) {\LARGE #3};}}
  \def\dense#1#2#3#4{
    \node[dense,#2] (d#10) {\mygrid{#2}{#3}{#4}}; 
    \&
    \node[dense,#2] (d#11) {\mygrid{#2}{#3}{#4}}; \&
    \node (d#12) {$\cdots$}; \&
    \node[dense,#2] (d#13) {\mygrid{#2}{#3}{#4}};
  }

  \def\dotsrow{
    \node[lbl] (s0) {$\vdots$}; \&
    \node[lbl] (s1) {$\vdots$}; \&
    \node[lbl] (s2) {$\vdots$}; \&
    \node[lbl] (s3) {$\vdots$};
  }
	\matrix[column sep={1cm,between origins},row sep=0.25cm,ampersand replacement=\&] (net) {
			\node[lbl] (l0) {\small SUB$_1$}; \&[-0.25cm]
			\sblock{embs}{embs.}{4}{400}{d10} \&[-0.075cm]
			\qrnn{1}{0}{5}{1550}{0.1cm}; \&
			\qrnn{2}{0}{5}{1550}{0.1cm}; \&
			\qrnn{3}{0}{5}{1550}{0.1cm}; \&[-0.075cm]
			\qrnn{4}{0}{4}{400}{0.1cm};\\

			\node[lbl] (l1) {\small SUB$_2$}; \&
			\sblock{embs}{embs.}{4}{400}{d11} \&
			\qrnn{1}{1}{5}{1550}{0.1cm}; \&
			\qrnn{2}{1}{5}{1550}{0.1cm}; \&
			\qrnn{3}{1}{5}{1550}{0.1cm}; \&
			\qrnn{4}{1}{4}{400}{0.1cm};\\

			\node (l2) {$\cvdots$};\&
			\node[text width=1cm,align=center] (d12) {$\cvdots$}; \&
			\node[text width=1.3cm,align=center] (r12) {$\cvdots$}; \&
			\node (r22) {$\cvdots$}; \&
			\node (r32) {$\cvdots$}; \&
			\node (r42) {$\cvdots$}; \\
			
			\node[lbl] (l3) {\small SUB$_n$}; \&
			\sblock{embs}{embs.}{4}{400}{d13} \&
			\qrnn{1}{3}{5}{1550}{0.1cm}; \&
			\qrnn{2}{3}{5}{1550}{0.1cm}; \&
			\qrnn{3}{3}{5}{1550}{0.1cm}; \&
			\qrnn{4}{3}{4}{400}{0.1cm};\\
	};

  \foreach \layer in {1,...,4}{
  	\draw[->] (r\layer0) -- (r\layer1);
  	\draw[->] (r\layer1) -- (r\layer2);
  	\draw[->] (r\layer2) -- (r\layer3);

  }

  \foreach \x in {0,1,3}{
    \draw[->] (l\x) -- (d1\x);
    \draw[->] (d1\x) -- (r1\x);
    \foreach \layer/\layerp in {1/2,2/3,3/4}{
    	\draw[->] (r\layer\x) -- (r\layerp\x);
    }
  }

  \draw[->] (d10.south east) -- (r11.north west);
  \draw[->] (d11.south east) -- (r12.north west);
  \draw[->] (d12.south east) -- (r13.north west);

  \matrix[column sep={0.7cm,between origins},row sep=0.25cm,ampersand replacement=\&,xshift=2.2cm] (head) at (net.east) {
  	\pool{avg}{avg pool}{4}{400} \&[0.1cm] \&\\
	\pool{max}{max pool}{4}{400} \& \pool{dense1}{linear$_1$}{3}{50} \&   \pool{dense2}{linear$_2$}{3}{\textit{\Large \#classes}}\\
	\pool{last}{last}{4}{400} \& \&\\
};
  \draw[-{Latex[width=0.5mm,length=0.5mm]},line width=0.25pt] (avg.east) -- (dense1.west);
  \draw[-{Latex[width=0.5mm,length=0.5mm]},line width=0.25pt] (max.east) -- (dense1.west);
  \draw[-{Latex[width=0.5mm,length=0.5mm]},line width=0.25pt] (last.east) -- (dense1.west);
  \draw[->] (dense1.east) -- (dense2.west);

  \foreach \x in {0,1,3}{
  	\draw[-{Latex[width=0.5mm,length=0.5mm]},line width=0.25pt] (r4\x.east) -- (avg.west);
  	\draw[-{Latex[width=0.5mm,length=0.5mm]},line width=0.25pt] (r4\x.east) -- (max.west);
  }
 \draw[-{Latex[width=0.5mm,length=0.5mm]},line width=0.25pt] (r43.east) -- (last.west); 
\end{tikzpicture}
\caption{The MultiFiT language model with classifier consisting of a subword embedding layer, four QRNN layers, an aggregation layer, and two linear layers.}
\label{fig:multifit}
\end{figure}

\begin{figure}
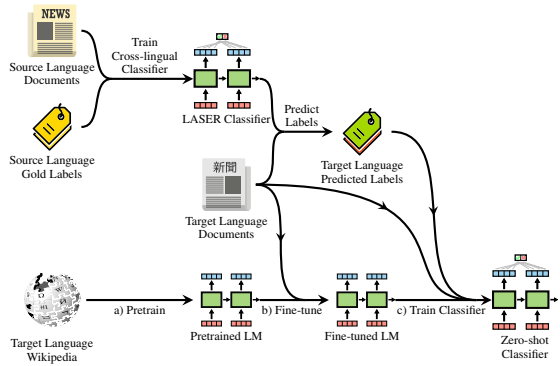

\centering
\subfile{gfx/pipeline}
\caption{The steps of our cross-lingual bootstrapping method for zero-shot cross-lingual transfer. a) A monolingual language model (LM) is pretrained on target language data; b) the LM is fine-tuned on target language documents; and c) the LM is fine-tuned as a classifier using the zero-shot predictions from a linear classification layer fine-tuned on top of cross-lingual representations from LASER.}
\label{fig:pipeline}
\end{figure}

\section{Our method}

\subsection{Multi-lingual Fine-Tuning}

We propose Multi-lingual Fine-tuning (MultiFit). Our method uses the ULMFiT model \cite{Howard2018} with discriminative fine-tuning as foundation. ULMFiT is based on a 3-layer AWD-LSTM \cite{Merity2017} language model. The AWD-LSTM is a regular LSTM \cite{Hochreiter1997} with tuned dropout hyper-parameters. To enable faster training and fine-tuning of the model, we replace it with a QRNN \cite{Bradbury2017}. The QRNN alternates convolutional layers, which are parallel across timesteps, and a recurrent pooling function, which is parallel across channels. It has been shown to outperform LSTMs, while being up to $16\times$ faster at train and test time. ULMFiT in addition is restricted to words as input. To make our model more robust across languages, we use subword tokenization based on a unigram language model \cite{Kudo2018}, which is more flexible compared to byte-pair encoding \cite{Sennrich2016}. We additionally employ label smoothing \cite{Szegedy2016} and a novel cosine variant of the one-cycle policy \cite{smith2018disciplined}\footnote{The idea is due to private conversation with the author.}, which we found to outperform ULMFiT's slanted triangular learning rate schedule and gradual unfreezing. The full model can be seen in Figure \ref{fig:multifit}.

\subsection{Cross-lingual Bootstrapping}

Prior methods have employed cross-lingual training strategies relying on parallel data and a shared BPE vocabulary. These can be combined with our language model, but increase its training complexity. For the case where an existing pretrained cross-lingual model and source language data are available, we propose a bootstrapping method \cite{Ruder2018} that uses the pretrained model's zero-shot predictions as pseudo labels to fine-tune the monolingual model on target language data.

The steps of the method can be seen in Figure \ref{fig:pipeline}. Specifically, we first fine-tune a linear classification layer on top of pretrained cross-lingual representations on source language training data. We then apply this cross-lingual classifier to the target language data and store its predicted label for every example. We now fine-tune our pretrained LM on the target language data and these pseudo labels\footnote{Distillation \cite{Hinton2015} yielded similar results.}. Importantly, this method enables our monolingual LM to significantly outperform its cross-lingual teacher in the zero-shot setting (\textsection \ref{sec:results}).

\section{Experimental setup}

This section provides an overview of our experimental setup; see the appendix for full details.

\paragraph{Data} We evaluate our models on the Multilingual Document Classification Corpus  \cite[MLDoc;][]{Schwenk2018}\footnote{\url{https://github.com/facebookresearch/MLDoc}}---a new subset of Reuters Corpus Volume 2 \cite{lewis2004rcv1} with balanced class priors for eight languages---and on the Cross-Lingual Sentiment dataset 
\cite[CLS;][]{Prettenhofer2010a}\footnote{\url{https://webis.de/data/webis-cls-10.html}} consisting of Amazon product reviews in four languages. We provide an overview of the datasets in Table \ref{tab:data}.

\begin{table}[]
\centering
\resizebox{\linewidth}{!}{%
\begin{tabular}{l l l r r r}
\toprule
 & Domain & Languages & Train & Dev & Test \\
\midrule
\multirow{2}{*}{MLDoc} & \multirow{2}{*}{News} & \en, \de, \es, \fr, & 1k / 2k / & \multirow{2}{*}{2k} & \multirow{2}{*}{10k}\\
& & \ita, \ja, \ru, \zh & 5k / 10k\\
\multirow{2}{*}{CLS} & Product & \multirow{2}{*}{\en, \de, \fr, \ja} & \multirow{2}{*}{2k} & \multirow{2}{*}{-} & \multirow{2}{*}{2k}\\
& reviews \\
\bottomrule
\end{tabular}%
}
\caption{The domain, languages, and number of training, development, and test examples in each dataset.}
\label{tab:data}
\end{table}

\paragraph{Pretraining} We pretrain our models on 100M tokens extracted from the Wikipedia of the corresponding language for 10 epochs. As fewer tokens might be available for some languages, we also compare against a version (no wiki) that uses no pretraining. For all models, we fine-tune the LMs on the target data of the same language for 20 epochs. We perform subword tokenization with the unigram language model \cite{Kudo2018}.

\begin{table}[]
\centering
\resizebox{\linewidth}{!}{%
\begin{tabular}{l c c c c c c c}
\toprule
                 & \de     & \es     & \fr     & \ita    & \ja     & \ru     & \zh \\
\midrule
\multicolumn{8}{l}{\emph{Zero-shot (1,000 source language examples)}} \\
\midrule
MultiCCA     &  81.20  &  72.50  &  72.38  &  69.38  &  67.63  &  60.80  &  74.73 \\
LASER, paper     &  86.25  &\B{79.30}&  78.30  &  70.20  &  60.95  &  67.25  &  70.98 \\
LASER, code      &  87.65  &  75.48  &  84.00  &  71.18  &  64.58  &  66.58  &  76.65 \\
MultiBERT        &  82.35  &  74.98  &  83.03  &  68.27  &  64.58  &  \B{71.58}  &  66.17\\
MultiFiT, pseudo &\B{91.62}&  79.10  &\B{89.42}&\B{76.02}&\B{69.57}& 67.83 &\B{82.48} \\
\midrule
\multicolumn{8}{l}{\emph{Supervised (100 target language examples)}} \\
\midrule
MultiFit        & 90.90 & 89.00 & 85.03 & 80.12 & 80.55 & 73.55 & 88.02 \\ 
\midrule
\multicolumn{8}{l}{\emph{Supervised (1,000 target language examples)}} \\
\midrule
MultiCCA   &  93.70  &  94.45  &  92.05  &  85.55  &  85.35  &  85.65  &  87.30 \\
LASER, paper    &  92.70  &  88.75  &  90.80  &  85.93  &  85.15  &  84.65  &  88.98 \\
MultiBERT       &  94.00  &  95.15  &  93.20  &  85.82  &  87.48  &  86.85  &  90.72 \\
Monolingual BERT & 94.93  &      -  &      -  &      -  &      -  &      -  &  92.17 \\
MultiFiT, no wiki & 95.23 & 95.07 & 94.65 & 89.30 & 88.63 & 87.52 & 90.03 \\
MultiFiT        &\B{95.90}&\B{96.07}&\B{94.75}&\B{90.25}&\B{90.03}&\B{87.65}&\B{92.52}	\\
\bottomrule
\end{tabular}%
}
\caption{Comparison of zero-shot and supervised methods on MLDoc.
}
\label{tab:mldoc-results}
\end{table}

\begin{table*}[]
\centering
\begin{tabular}{l l | lll | lll | lll}
\toprule
&  & \multicolumn{3}{c|}{\de} & \multicolumn{3}{c|}{\fr} & \multicolumn{3}{c}{\ja}\\
&  & Books & DVD & Music & Books & DVD & Music & Books & DVD & Music\\
\midrule
\multirow{3}{*}{\rotatebox[origin=c]{90}{\emph{Zero-shot}}} & LASER, code & 84.15 & 78.00 & 79.15 & 83.90 & 83.40 & 80.75 & 74.99 & 74.55 & 76.30 \\
 & MultiBERT  & 72.15 & 70.05 & 73.80 & 75.50 & 74.70 & 76.05 & 65.41 & 64.90 & 70.33 \\
 & MultiFiT, pseudo & \B{89.60} & \B{81.80} & \B{84.40} & \B{87.84} & \B{83.50} & \B{85.60} & \B{80.45} & \B{77.65} & \B{81.50} \\
\midrule
\multirow{3}{*}{\rotatebox[origin=c]{90}{\emph{Translat.}}} & MT-BOW & 79.68 & 77.92 & 77.22 & 80.76 & 78.83 & 75.78 & 70.22 & 71.30 & 72.02\\
& CL-SCL & 79.50 & 76.92 & 77.79 & 78.49 & 78.80 & 77.92 & 73.09 & 71.07 & 75.11 \\
& BiDRL & 84.14 & 84.05 & 84.67 & 84.39 & 83.60 & 82.52 & 73.15 & 76.78 & 78.77 \\
\midrule
\multirow{2}{*}{\rotatebox[origin=c]{90}{\emph{Super.}}} & MultiBERT  & 86.05 & 84.90 & 82.00 & 86.15 & 86.90 & 86.65 & 80.87 & 82.83 & 79.95 \\
 & MultiFiT & \textbf{93.19} & \textbf{90.54} & \textbf{93.00} & \textbf{91.25} & \textbf{89.55} & \textbf{93.40} & \textbf{86.29} & \textbf{85.75} & \textbf{86.59} \\
\bottomrule
\end{tabular}%
\caption{Comparison of zero-shot, translation-based and supervised methods (with 2k training examples) on all domains of CLS. MT-BOW and CL-SCL results are from \cite{Zhou2016c}. }
\label{tab:cls-complete}
\end{table*}

\paragraph{Evaluation settings} We compare two settings based on the availability of source and target language data: supervised and zero-shot. In the supervised setting, every model is fine-tuned and evaluated on examples from the target language. In the zero-shot setting, every model is fine-tuned on source language examples and evaluated on target language examples. In all cases, we use English as the source language.

\paragraph{Baselines} We compare against the state-of-the-art cross-lingual embedding models LASER \cite{Artetxe2018e}, which uses a large parallel corpus, multilingual BERT (MultiBERT)\footnote{\url{https://github.com/google-research/bert/blob/master/multilingual.md}}, and monolingual BERT\footnote{Models are available for English, Chinese, and German (\url{https://deepset.ai/german-bert}).}. We also compare against the best models on each dataset, MultiCCA \cite{Ammar2016a}, a cross-lingual word embedding model, and BiDRL \cite{Zhou2016c}, which translates source and target data.

\paragraph{Our methods} We evaluate our monolingual LMs in the supervised setting (MultiFit) and our LMs fine-tuned with pseudo labels from LASER in the zero-shot setting (pseudo).

\begin{table}[]
\centering
\resizebox{0.8\linewidth}{!}{%
\begin{tabular}{l c c}
\toprule
& LSTM & QRNN \\
\midrule
Language model pretraining & 143 & \textbf{71} \\
Classifier fine-tuning & 467 & \textbf{156} \\
\bottomrule
\end{tabular}%
}
\caption{Comparison of LSTM and QRNN per-batch training speed on a Tesla V100 (in ms) in MultiFiT.}
\label{tab:speed}
\end{table}

\begin{table}[h!]
\centering
\resizebox{\linewidth}{!}{%
\begin{tabular}{l c c c}
\toprule
                        & \de     & \es & \zh \\
\midrule
ULMFiT                & 94.19   & 95.23 &    66.82   \\
MultiFiT, no wiki      & 95.23   & 95.07  &   90.03 	 \\
MultiFiT, small Wiki   & 95.37   & 95.30   &  89.80	 \\
MultiFiT              &\B{95.90}&\B{96.07}&\B{92.52} \\
\bottomrule
\end{tabular}%
}
\caption{Comparison of MultiFiT results with different pretraining corpora and ULMFiT, fine-tuned with 1k labels on MLDoc.}
\label{tab:ulmfit-results}
\end{table}

\section{Results} \label{sec:results}


\paragraph{MLDoc} We show results for MLDoc in Table \ref{tab:mldoc-results}. In the zero-shot setting, MultiBERT underperforms the comparison methods as the shared embedding space between many languages is overly restrictive. Our monolingual LMs outperform their cross-lingual teacher LASER in almost every setting. When fine-tuned with only 100 target language examples, they are able to outperform all zero-shot approaches except MultiFiT on \de{} and \fr. This calls into question the need for zero-shot approaches, as fine-tuning with even a small number of target examples is able to yield superior performance. When fine-tuning with 1,000 target examples, MultiFiT---even without pretraining---outperforms all comparison methods, including monolingual BERT.

\paragraph{CLS} We show results for CLS in Table \ref{tab:cls-complete}. MultiFiT is able to outperform its zero-shot teacher LASER across all domains. Importantly, the bootstrapped monolingual model also outperforms more sophisticated models that are trained on translations across almost all domains. In the supervised setting, MultiFiT similarly outperforms multilingual BERT. For both datasets, our methods that have been pretrained on 100 million tokens outperform both multilingual BERT and LASER, models that have been trained with orders of magnitude more data and compute.


\section{Analysis}

\paragraph{Speed} We compare the LSTM and QRNN cell in MultiFiT based on the speed for processing a single batch for pretraining and fine-tuning in Table \ref{tab:speed}. MultiFiT with a QRNN pretrains and fine-tunes about $2\times$ and $3\times$ faster respectively. 

\paragraph{MultiFiT vs. ULMFiT} We compare MultiFiT pretrained on 100M Wikipedia tokens against ULMFiT pretrained on the same data using a 3-layer LSTM and spaCy tokenization\footnote{\url{https://spacy.io/api/tokenizer}} as well as MultiFiT pretrained on 2M Wikipedia tokens, and MultiFiT with no pretraining in Table \ref{tab:ulmfit-results}. Pretraining on more data generally helps. MultiFiT outperforms ULMFiT significantly; the performance improvement is particularly pronounced in Chinese where ULMFiT's word-based tokenization underperformed.

\begin{figure}[h!]
\includegraphics[width=8cm]{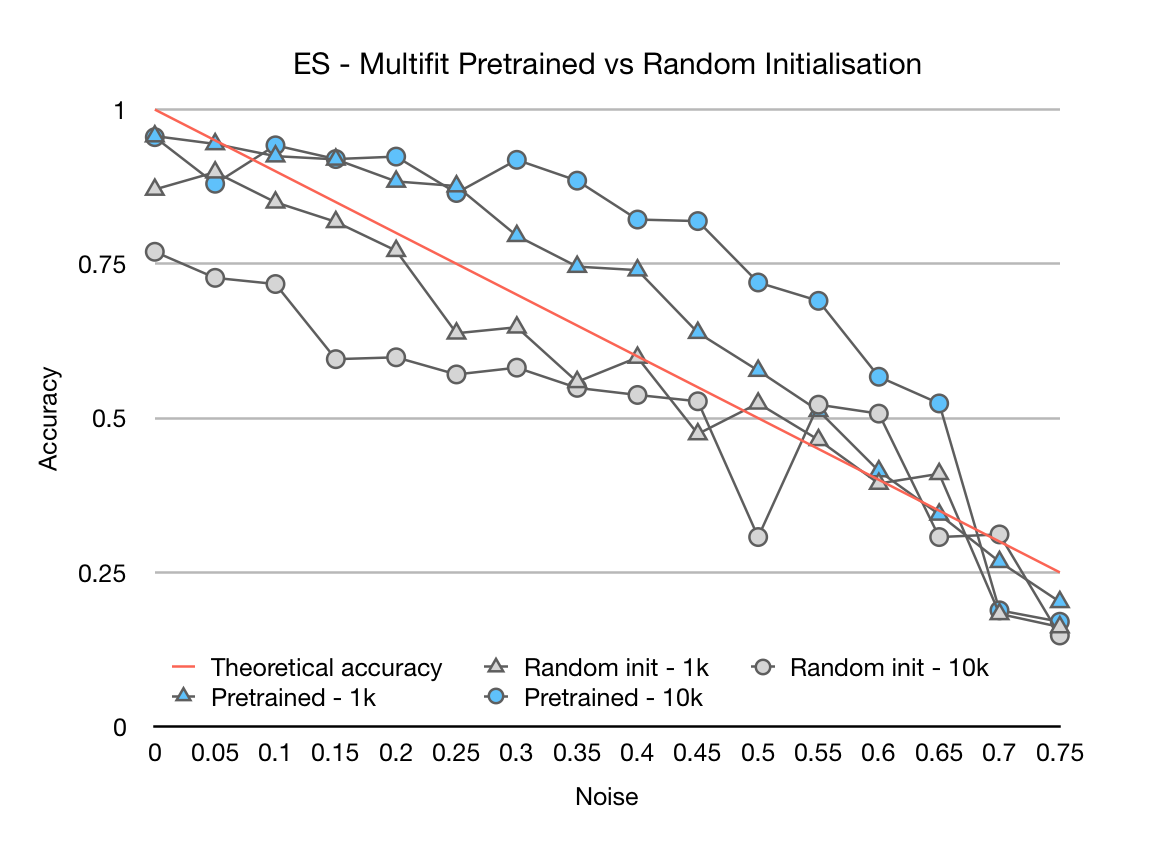}
\caption{Comparison of MultiFiT's robustness to label noise on MLDoc with and without pretraining.}
\label{fig:noise}
\end{figure}

\begin{table}[h!]
\centering
\resizebox{\linewidth}{!}{%
\begin{tabular}{l c c c c c c c}
\toprule
                   & \de     & \es     & \fr     & \ita    & \ja     & \ru     & \zh \\
\midrule
LASER, code &  87.65  &  75.48  &  84.00  &  71.18  &  64.58  &  66.58  &  76.65 \\
\midrule
Random init. (1k)  &  77.80  &  70.50  &  75.65  &  68.52  &  68.50  &  61.37  &  79.19 \\
Random init. (10k) &  90.53  &  69.75  &  87.40  &  72.72  &  67.55  &  63.67  &  81.44 \\
MultiFiT, pseudo (1k)          &\B{91.34}&\B{78.92}&\B{89.45}&\B{76.00}&\B{69.57}&\B{68.19}&\B{82.45} \\
\bottomrule
\end{tabular}%
}
\caption{Bootstrapping results on MLDoc with and without pretraining, trained on 1k/10k LASER labels.}
\label{tab:rnd-results}
\end{table}



\begin{table}[h!]
\centering
\resizebox{\linewidth}{!}{%
\begin{tabular}{l c c c c c}
\toprule
& \de & \es & \fr & \ita & \ru \\
\midrule
Word-based & 95.28 & 95.97 & 94.72 & 89.97 & \textbf{88.02} \\
Subword & \textbf{96.10} & \textbf{96.07} & \textbf{94.75} & \textbf{94.75} & 87.65\\
\bottomrule
\end{tabular}%
}
\caption{Comparison of different tokenization strategies for different languages on MLDoc.}
\label{tab:tok}
\end{table}


\paragraph{Robustness to noise} We suspect that MultiFiT is able to outperform its teacher as the information from pretraining makes it robust to label noise. To test this hypothesis, we train MultiFiT and a randomly initialized model with the same architecture on 1k and 10k examples of the Spanish MLDoc. We randomly perturb labels with a probability ranging from 0-0.75 and show results in Figure \ref{fig:noise}. The pretrained MultiFiT is able to partially ignore the noise, up to 65\% of noisy training examples. Without pretraining, the model does not exceed the theoretical baseline (the percentage of correct examples). In addition, we compare MultiFiT with and without pretraining in Table \ref{tab:rnd-results}. Pretraining enables MultiFiT to achieve much better performance compared to a randomly initialised model. Both results together suggest a) that pretraining increases robustness to noise and b) that information from monolingual and cross-lingual language models is complementary.


\paragraph{Tokenization} Subword tokenization has been found useful for language modeling with morphologically rich languages \cite{Czapla2018,Mielke2019} and has been used in recent pretrained LMs \cite{Devlin2018}, but its concrete impact on downstream performance has not been observed. We train models with the best performing vocabulary sizes for subword (15k) and regular word-based tokenization (60k) with the Moses tokenizer \cite{koehn2007moses} on German and Russian MLDoc and show results in Table \ref{tab:tok}. Subword tokenization outperforms word-based tokenization on most languages, while being faster to train due to the smaller vocabulary size.

\section{Conclusion}

We have proposed novel methods for multilingual fine-tuning of languages that outperform models trained with far more data and compute on two widely studied cross-lingual text classification datasets on eight languages in both the zero-shot and supervised setting.

\bibliography{acl2019}
\bibliographystyle{acl_natbib}

\newpage
\newpage

\appendix

\section{Hyper-parameters}
The MultiFit architecture has 4 QRNN layers with a hidden dimensionality of 1550, a vocabulary size of 15,000 subword tokens, and an embeding size of 400. The vocabularies were computed using the SentencePiece\footnote{\url{https://github.com/google/sentencepiece}} unigram language model~\cite{Kudo2018} with 99\% character coverage for Chinese and Japanese and 100\% for the rest. The encoder's and decoder's weights are shared~\cite{Press2017}. The output of the last QRNN layer (the last time step concatenated with an average and maximum pooled over time steps) is passed to the classifier with 2 dense layers.

Our language models were trained for 10 epochs on 100 million tokens of Wikipedia articles and then fine-tuned for 20 epochs on the corresponding dataset (MLDoc or CLS). The classifier was fine-tuned for 4 to 8 epochs. Results of the best model based on accuracy on the validation set are reported. We used a modified version of 1-cycle learning rate schedule \cite{smith2018disciplined} that uses cosine instead of linear annealing, cyclical momentum and discriminative finetuning \cite{Howard2018}. Our batch size for language model training was 50 and for classification tasks 18. We were using BPTT of length 70. Due to the large amount of available training data our pretrained language models were trained without any dropout. We used the same dropout values as \cite{Howard2018} multiplied by 0.3 and 0.5 for fine-tuning of language models and the classification task respectively. We used weight decay of 0.01 for both tasks. The final regularization method was label smoothing \cite{Szegedy2016} with epsilon of $0.1$.

\section{Speed comparison hyper-parameters}

For the speed comparison, we use the same architecture and only change the underlying RNN cell (QRNN or LSTM). We pretrain and fine-tune both models on 15k tokens on a Tesla V100. For pretraining, we use a BPTT size of 70 and a batch size of 64. For fine-tuning, we use a batch size of 32.

\end{document}